\documentclass[runningheads]{llncs}
\usepackage[T1]{fontenc}
\usepackage{graphicx}
\usepackage{amssymb}
\usepackage{amsmath}
\usepackage{multirow}
\usepackage{arydshln}
\usepackage{subcaption}
\usepackage{xcolor}

\begin{document}

\title{Empirical Evaluation of Out-Of-Distribution Performance of Tabular Foundation Models}

\titlerunning{Empirical Evaluation of Out-of-Distribution Performance of Tabular}
%

\author{
Malena Loza\inst{1} \and 
David Chushig-Muzo\inst{2} \and
Eva Milara\inst{2} \and 
Luis Bote-Curiel\inst{2} \and 
Luis Estrada-Petrocelli\inst{3} \and 
Felipe Grijalva\inst{1}
}

\authorrunning{Loza-Casa M. et al.}

\institute{
Colegio de Ciencias e Ingenierías, Universidad San Francisco de Quito (USFQ), Quito, Ecuador; \email{\{mloza,fgrijalva\}@usfq.edu.ec} 
\and
Department of Signal Theory and Communications, Telematics and Computing Systems, Rey Juan Carlos University, Madrid, Spain; \email{\{david.chushig,eva.milara,luis.bote\}@urjc.es}
\and
Facultad de Ingeniería, Universidad Latina de Panamá, Ciudad de Panamá, Panamá; \email{lestrada@ulatina.edu.pa}
}

\maketitle 

\begin{abstract}
Tabular Foundation Models (TFMs) have emerged as novel approaches for tabular predictive tasks, demonstrating competitive predictive performance to ensemble tree-based models. Most TFMs are trained and evaluated on independent and identically distributed data, but this assumption changes in real-world scenarios due to distribution shifts, which compromise the robustness of models. Limited research has been conducted of TFMs under distribution shifts. We present an empirical evaluation of Out-Of-Distribution (OOD) performance of nine TFMs, spanning diverse pre-training strategies and architectures: TabPFNv2, TabPFNv2.5, TabPFNv2.6, TabPFNv3, TabICL, TabICLv2, Mitra, LimiX and TabFM. Three real-world datasets from the TableShift study were considered (HELOC, Voting, Childhood Lead), covering label, socioeconomic, and geographic shift types. Our results show that all evaluated TFMs degrade systematically under distribution shift regardless of pre-training strategy, with shift gaps ranging from 0.003 to 0.060 depending on shift type. The relationship between in-distribution and OOD predictive performance documented for classical tabular models extends into TFMs. We also identified a scalability gap, as high-performing models demand significant memory and computational resources beyond what standard deployment infrastructure can support. This study extends existing benchmarks for OOD in tabular data, providing evidence to support their adoption in high-stakes domains characterized by structural distribution shifts.
\end{abstract}

\section{Introduction}

Tabular data are among most extended data formats for representing structured information, which is characterized by a \textit{table-like format}, where rows and columns correspond to samples and features, respectively~\cite{jiang2026representation}. For years, tree-based ensemble models, including XGBoost, LightGBM, and CatBoost, have been the state-of-the-art in predictive tabular tasks due to their high performance and interpretability~\cite{somvanshi2026survey}. Although several attempts of using deep learning models for tabular data have been proposed (\textit{e.g.,} TabNet~\cite{arik2021tabnet}, SAINT~\cite{somepalli2021saint}), they have shown lower performance compared to tree-based models.

Tabular Foundation Models (TFMs) have emerged as novel approaches, showing remarkable generalization to downstream tasks without fine-tuning~\cite{jiang2026representation}. A remarkable type of TFMs, often referred to as Prior-Data Fitted Networks (PFNs), exploits in-context learning (ICL) at inference time, passing the labeled training subset as context and predicting test samples without gradient updates. These TFMs have been trained using diverse collections of tabular data, including synthetic and real-world datasets, and or mixtures of both~\cite{somvanshi2026survey}. In particular, TabPFN marked a milestone for tabular classification, outperforming tree-based ensemble models on small data~\cite{hollmann2025accurate}. Recent TFMs such as TabDPT~\cite{ma2024tabdpt}, Mitra~\cite{zhang2026mitra}, TabICL~\cite{qu2026tabiclv2}, and Limix~\cite{zhang2025limix}, have also outperformed tree-based models across different benchmarks. TFMs have been used in a range of applications such as disease classification~\cite{chushig2024characterizing,lara2025interpretable}, intrusion detection~\cite{ruiz2025wfe,ruiz2024tabpfn} among others. 

Despite the success of TFMs, existing models are developed under the assumption that all samples of training and test subsets follow the same In-Distribution (ID). However, this assumption often breaks down in real-world applications, where deployed models are often exposed to unseen samples drawn from distributions that differ from the training data, known as Out-Of-Distribution (OOD). Such distributional shifts can significantly degrade model performance, resulting in inaccurate predictions of underrepresented domains~\cite{yang2024generalized}. While existing benchmarks (such as TableShift~\cite{gardner2023benchmarking}) have evaluated the OOD effect on traditional machine learning models, the robustness of TFMs to distribution shifts remains underexplored. 

In this paper, we conduct an empirical evaluation of the performance of TFMs under OOD scenarios. We evaluate nine TFMs with different pre-training strategies, including: \textit{(i)} real-world data pre-training (TabPFN v2, TabPFN v2.5, TabPFN v2.6, TabPFN v3, TabFM); \textit{(ii)} mixed prior structural causal models with tree ensemble priors (Mitra); and \textit{(iii)} synthetic priors (TabICL, TabICL v2, LimiX). We evaluate on three TableShift datasets, covering label shift (HELOC), socioeconomic shift (Childhood Lead), and geographic shift (Voting). These were selected to cover three application domains (healthcare, finance and public health). To our knowledge, this is the first work to investigate the impact of OOD and distribution shifts on the predictive performance of TFMs.

The main contributions of this study are:
\begin{itemize}
    \item We provide the first empirical evaluation of TFMs under distribution shifts, covering three real-world datasets, and three shift types.
    \item We evaluate the impact of pre-training approach (with synthetic, real-world, or mixed data) of TFMs on OOD robustness.
    \item We extend the OOD analysis of tabular predictive models into TFMs, assessing the relationship between ID and OOD performance.
    \item We provide evidence of the computational constraints of TFMs under dataset sizes, finding a practical implication for model deployment.
    \item We provide empirical evidence of TFMs under distribution shifts across healthcare and financial domains.
\end{itemize}

\section{Materials and methods}

\subsection{Datasets}

We evaluate different TFMs on three datasets from TableShift~\cite{gardner2023benchmarking}, selected to cover a range of domains and distribution shift types: HELOC, Childhood Lead, and Voting. Table~\ref{table:summary_datasets} summarizes the datasets considered. HELOC is a financial risk dataset, where the task is to predict whether a consumer will repay a home equity line of credit. The distribution shift in this dataset is characterized by label shift, \textit{e.g.,} the marginal distribution of the target variable differs substantially between the ID and OOD subsets. Childhood Lead is derived from the National Health and Nutrition Examination Survey. The task is to predict whether a child has elevated blood lead levels. The domain split is defined by household poverty level, introducing a socioeconomic subpopulation shift between ID and OOD. 

\begin{table}
\centering
\caption{A summary of datasets used in this study.}
\label{table:summary_datasets}
\begin{tabular}{|l|l|l|l|l|l|l|}
\hline
\textbf{Dataset} & \textbf{Shift Variable} & \textbf{Features} & \textbf{Train} & \textbf{Validation} & \textbf{ID test} & \textbf{OOD test} \\ \hline
Voting & Geographic region  & 380 & 37548 & 4693 & 4694 & 23103 \\
Childhood lead & Poverty level      & 17  & 11807 & 1476 & 1476 & 11466 \\
HELOC & Label distribution  & 38  & 2220  & 278  & 278  & 6914 \\
\hline
\end{tabular}
\end{table}

Voting is constructed from the American National Election Studies Time Series Cumulative Data File, covering presidential elections from 1948 to 2020. The task is to predict voter turnout. The domain split is defined by U.S. Census region, with the Southern states constituting the OOD subset and other regions forming the training distribution. This geographic shift reflects a real-world scenario in which a model trained on one regional population is applied to another with different political and demographic characteristics.

\subsection{Tabular foundation models}

TFMs comprise a broad group of approaches for transferring knowledge across tabular prediction tasks. A prominent line within this group is formed by PFNs and related ICL-based TFMs, which perform inference without the need for parameter updates and provide competitive predictive performance on small-scale datasets. TFMs present differences in pre-training approaches. Some models are pre-trained on collections of real-world tabular datasets, aiming to learn statistical patterns that generalize across different domains. In contrast, other models are pre-trained exclusively on synthetic datasets generated from Structural Causal Models (SCMs) (or probabilistic generative processes), with no exposure to real-world tabular data.


TabPFN v1 uses a transformer architecture to perform ICL on embeddings of rows. It was pre-trained on millions of  datasets generated from SCMs to approximate Bayesian inference over tabular classifiers~\cite{hollmann2025accurate}. This enables the model to learn transferable causal inductive biases that generalize to unseen real-world data. The original TabPFN was designed for datasets with a maximum context window of 1,000 training rows and 100 numerical features, exhibiting limited performance on categorical data~\cite{behre2026context}. To address scalability, several variants have been proposed. TabPFN v2 includes dense token subsampling and improved task conditioning, supporting up to 10,000 samples~\cite{lee2026multitabpfn}. TabPFN v2, v2.5, and v2.6 uses a transformer architecture that alternates row-wise and feature-wise attention, enhancing predictive performance but introducing computational complexity that scales poorly with dataset size~\cite{grinsztajn2026tabpfn}.

TabPFN v3 introduces several changes in the architecture to improve scalability. For classification, it employs an attention-based retrieval decoder instead of a fixed-width multilayer perceptron output head, allowing it to handle an arbitrary number of classes via soft nearest-neighbor retrieval~\cite{grinsztajn2026tabpfn}. It introduces row chunking, enabling fixed-size row processing that decouples GPU memory usage from dataset size while preserving the predictions of full inference. Also, it leverages multi-query attention to minimize key–value cache memory consumption during inference~\cite{grinsztajn2026tabpfn}. 

TabFM is a zero-shot ICL-based TFM developed by Google Research that presents the strengths of both TabPFN and TabICL using a novel architecture built with three mechanisms: \textit{(i)} alternating row-and-column attention by capturing complex feature interactions without manual feature engineering; \textit{(ii)} row compression that condenses each row's cross-attended representation into a single dense vector; and \textit{(iii)} a dedicated ICL transformer that operates over compressed row embeddings. In contrast to the TabPFN variants, TabFM is trained on hundreds of millions of synthetic datasets generated using SCMs. 

TabICL addresses the scalability bottleneck of TFMs by replacing the joint attention with a two-stage architecture, reducing complexity and scaling to large number of samples~\cite{qu2026tabiclv2}. Pre-trained on synthetic data generated from SCMs, TabICL achieves competitive performance while requiring less memory and lower computational resources than TabPFN variants. TabICLv2 introduces three key improvements over TabICL: \textit{(i)} a novel synthetic data generation engine for high pretraining diversity; \textit{(ii)} architectural enhancements (including a scalable softmax); and \textit{(iii)} a new pre-training approach by replacing AdamW with the Muon optimizer~\cite{qu2026tabiclv2}. 

Mitra demonstrates that mixed priors combining SCMs and tree ensembles effectively model decision boundary geometry~\cite{zhang2026mitra}. In contrast to TabPFNv2 and TabICLv2, which merge each feature and its label into a single token, Mitra represents the label using a separate token for each row. LimiX represents tabular data as a set-valued sample–feature entity rather than collapsing each row into a fixed embedding, enabling it to model dependencies across both samples and features~\cite{zhang2025limix}. 

\section{Results}
\label{sec:results}

\subsection{Experimental setup}

TFMs are used in their default inference configurations without task-specific fine-tuning, and employing the default setup in the original papers. For each dataset we use the three splits defined by TableShift: a training split used as the in-context examples for TFM inference, an ID test split drawn from the same distribution as training, and an OOD test split drawn from the held-out domain. No validation data is used for hyperparameter tuning, as all TFMs are evaluated in their default configurations.

We report the Area Under the Receiver Operating Characteristic (ROC-AUC) values for each model-dataset combination. Due to the class imbalance present in several datasets, ROC-AUC is considered as the evaluation metric. The shift gap ($\Delta$ ROC-AUC) is defined for each metric as the difference between ID and OOD performance, where larger positive values indicate greater degradation under shift. All experiments are repeated across five random seeds. We report mean and standard deviation across seeds. Due to the high inference computational cost of TabFM and Limix, combined with the size of the Voting dataset, results for this dataset are not reported.

\subsection{Classification results}

\begin{figure}[!htbp]
    \centering
  \begin{subfigure}[b]{0.99\linewidth}
    \includegraphics[width=\linewidth]{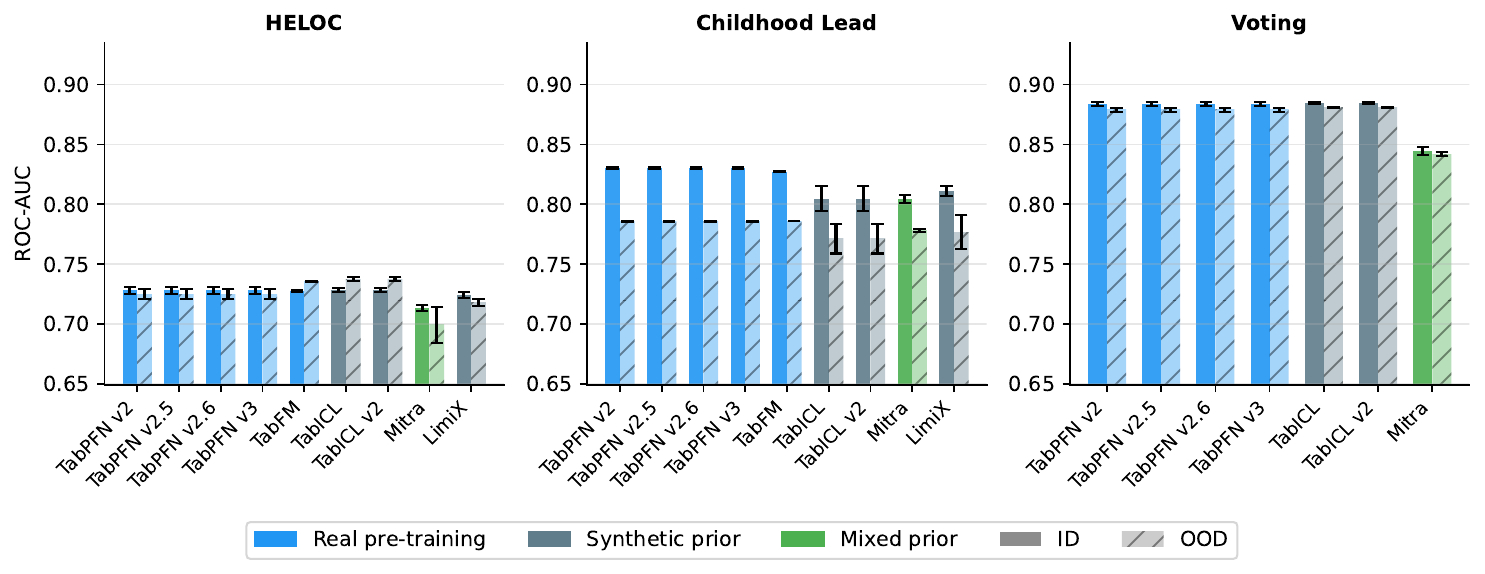} 
  \end{subfigure}
  \captionsetup{justification=justified, singlelinecheck=false, margin=6pt} 
  \caption{In-distribution and out-of-distribution results for all tabular foundation models across the datasets: HELOC, childhood lead, and voting.}
  \label{fig:id_ood_results} 
\end{figure}

Figure~\ref{fig:id_ood_results} presents the ID and OOD ROC-AUC for all TFMs and the datasets considered. OOD performance was consistently lower than ID performance, confirming that distribution shift degrades predictive results for all TFMs. Models based on real-world pre-training (TabPFN variants) consistently achieved the highest performance both ID and OOD across all three datasets. The shift gap varies substantially by dataset and shift type: \textit{(i)} Childhood lead (socioeconomic shift) showed the largest degradation, with all models exhibiting a drop from ID to OOD; \textit{(ii)} HELOC (label shift) showed a more moderate gap; and \textit{(iii)} Voting (geographic shift) presented the smallest degradation. Mitra performed competitively on Voting dataset, but showed the largest degradation on HELOC and  Childhood Lead, being less robust to distribution shifts. The gap between ID and OOD persists across all pre-training strategies and all shift types evaluated.

\begin{figure}[!htbp]
    \centering
  \begin{subfigure}[b]{0.99\linewidth}
    \includegraphics[width=\linewidth]{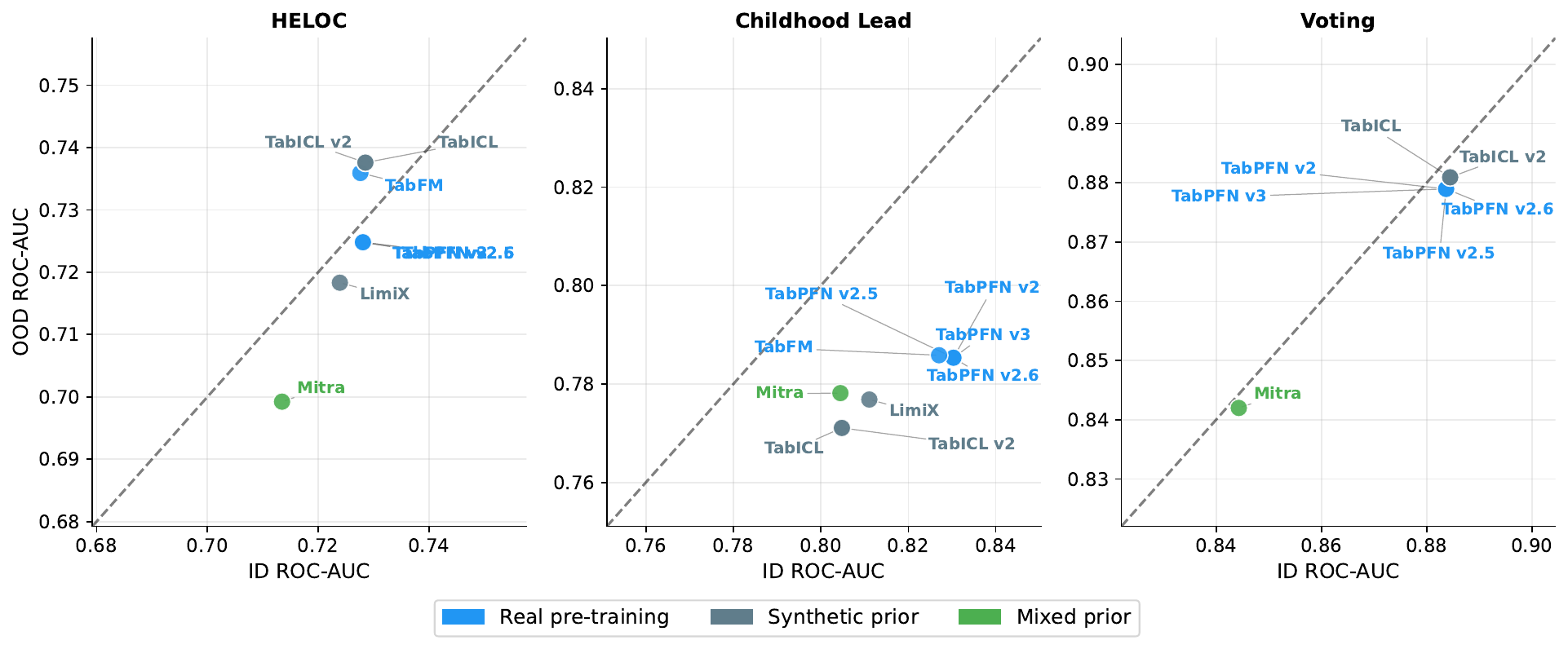} 
  \end{subfigure}
  \captionsetup{justification=justified, singlelinecheck=false, margin=6pt} 
  \caption{In-distribution versus out-of-distribution ROC-AUC scatter plots per dataset. Each point represents one model, coloured by pre-training strategy (blue: real-world, slate: synthetic, green: mixed). The dashed diagonal line represents no degradation under shift, points below the diagonal indicate OOD performance worse than ID.}
  \label{fig:scatter_id_ood} 
\end{figure}

Figure~\ref{fig:scatter_id_ood} shows the ID versus OOD ROC-AUC values for each model across the three datasets, with the dashed diagonal representing the no-degradation reference. All points lying below the diagonal confirm that every model suffers performance degradation under distribution shift, with no model achieving OOD parity with its ID performance across any dataset. On HELOC (label shift), models place in a narrow ID range (0.70–0.74) but diverge more on OOD performance: TabICL and TabICL v2 achieved the highest OOD scores despite moderate ID performance, while Mitra shows the largest drop, indicating high sensitivity to label shift. LimiX also falls below other models in OOD performance relative to its ID ROC-AUC value. 

On Childhood Lead (socioeconomic shift), the TabPFN variants (v2, v3) occupies the top-right region (highest ID and OOD), while synthetic-prior models (TabICL, TabICL v2) place at lower ID performance and suffer a larger gap from the diagonal. Mitra and LimiX again fall below most models in OOD performance. On Voting (geographic shift), all models except Mitra achieved high ID performance (0.88–0.90) and the shift gap is visually small, with most points lying close to the diagonal. TabICL and TabICL v2 are competitive with the TabPFN variants. Across all three panels, the distance below the diagonal is largest for Childhood Lead, and smallest for Voting and HELOC, reinforcing that label shift is the most relevant shift type for TFMs, consistent with findings from the original TableShift study on traditional models.

\begin{figure}[!htbp]
    \centering
  \begin{subfigure}[b]{0.99\linewidth}
    \includegraphics[width=\linewidth]{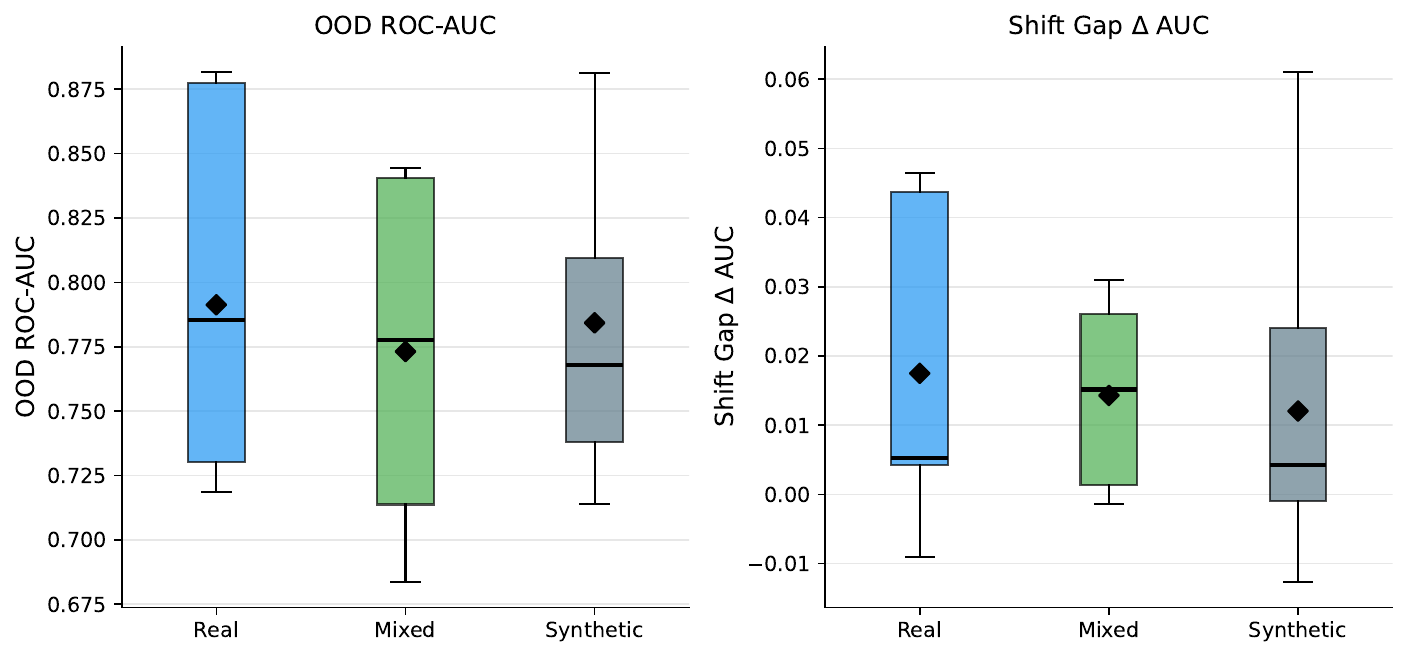} 
  \end{subfigure}
  \captionsetup{justification=justified, singlelinecheck=false, margin=6pt} 
  \caption{Pre-training corpus effect on out-of-distribution robustness. Left panel: distribution of out-of-distribution ROC-AUC by pre-training strategy. Right panel: distribution of shift gap  ROC-AUC. }
  \label{fig:boxplots} 
\end{figure}

Figure~\ref{fig:boxplots} compares OOD ROC-AUC and Shift Gap $\Delta$ across the three pre-training corpus strategies: Real, Mixed, and Synthetic. The left panel revealed that real-data pre-training achieved the highest median and mean OOD performance (median $\approx$ 0.78, mean $\approx$ 0.79), followed by Mixed (median $\approx$ 0.775) and Synthetic (median $\approx$ 0.77), though the differences in central tendency are modest. Notably, the Real group exhibits the widest interquartile range, reflecting the diversity of TabPFN variants (v2 through v3) and TabFM, which span a broader performance spectrum across datasets. 

In the right panel, the Real group shows the largest mean shift gap (mean $\approx$ 0.019), indicating that real pre-training models can suffer substantial degradation under certain shift types. The Mixed and Synthetic groups both exhibit lower mean shift gaps (mean $\approx$ 0.015 and $\approx$ 0.012, respectively), with Synthetic showing the lowest median gap of the three groups. Both Mixed and Synthetic include negative gap values, meaning some models in these groups marginally improve under OOD conditions on certain dataset. These findings suggest that pre-training corpus type is not the primary factor of OOD robustness. While real pre-training models tend to achieve higher absolute OOD performance, they do not exhibit smaller shift gaps, indicating that the performance advantage observed in Figure~\ref{fig:id_ood_results} is driven by stronger ID performance rather than by superior distributional robustness. 

\section{Discussion}

This study presented a comprehensive evaluation of TFMs under distribution shifts, covering nine TFMs across three real-world TableShift datasets spanning label shift (HELOC), socioeconomic shift (Childhood Lead), and geographic shift (Voting). Our results yield five main findings.

First, pre-training with real-world datasets does not provide OOD robustness. While models pre-trained on real-world data (TabPFN variants, TabFM) achieved higher absolute OOD performance, their shift gaps are comparable to synthetic-prior models (TabICL, LimiX). These results suggest that real-data pre-training does not improve robustness to distribution shifts. Second, it was revealed that relationship between ID and OOD performance documented by Gardner et al.~\cite{gardner2023benchmarking} for traditional tabular models extends to TFMs. In all three datasets, models with higher ID performance tend to achieve higher OOD performance, with no model consistently breaking above or below the proportional degradation line. The results indicate that improvements in ID performance, achieved through improved pre-training, larger context windows, or architectural refinements, are reflected proportionally under distribution shift. However, higher ID performance does not necessarily result in better OOD generalization, indicating that TFMs lack an inherent robustness advantage.

Third, the type of distribution shift was the main factor of OOD degradation. Across all models, Childhood Lead (socioeconomic subpopulation shift) produced the largest performance losses, HELOC (label shift) showed intermediate degradation, and Voting (geographic shift) exhibited the smallest gap. This ordering aligns with findings from TableShift, which identify label shift as one of the main factors associated with OOD degradation. This pattern was consistent across pre-training strategies, with no model exhibiting a systematic advantage under distribution shifts, suggesting that the mechanisms driving performance degradation are independent of the pre-training prior.

Fourth, Mitra demonstrates greater sensitivity to label shift than the other evaluated models. Although its performance remains competitive under socioeconomic and geographic shifts, it exhibits the largest OOD degradation on HELOC. This pattern suggests that its pre-training prior, which combines SCMs with tree ensembles, is well suited to covariate and subpopulation shifts but less robust to changes in the label distribution. These findings suggest that Mitra should be used with caution in environments where label prevalence may differ between training and deployment, such as clinical risk prediction across healthcare systems with different patient populations.

Fifth, scalability remains a major practical limitation of TFMs. A recurring challenge was the inability of several TFMs to process large datasets without significant computational constraints. In particular, TabFM and LimiX exhibited high memory growth with respect to the number of training samples and features, making them impractical on standard hardware for datasets with medium size. LimiX's cross-attention mechanism scales as $\mathcal{O}(n_{\text{train}} \times n_{\text{test}} \times d_{\text{model}})$, requiring chunked inference and training context subsampling even on a 47GB GPU (NVIDIA RTX 6000 Ada) for the Childhood Lead dataset (11,807 training rows, 17 features) and making experiments with Voting dataset (37,548 rows, 380 features) infeasible without subsampling. Similarly, TabFM required chunkinf on OOD test sets to avoid out-of-memory errors, despite being explicitly designed for scalability. Mitra imposed a hard limit of 10,000 training rows due to quadratic memory growth, a constraint that affected three of the five datasets considered in this study. These results reveal a fundamental trade-off: the highest-performing TFMs are often the least scalable, limiting their deployment to environments with high-end GPU resources and specialized memory-management techniques. Existing benchmarks largely obscure this limitation because they evaluate models on relatively small datasets that remain well within the memory capacity of all methods.

\section{Conclusions}

This paper presented the first OOD evaluation of modern TFMs, evaluating nine TFMs spanning synthetic, mixed, and real-world pre-training strategies. We evaluated on three TableShift datasets covering label, socioeconomic, and geographic shifts. Our results showed that TFMs do not exhibit inherent robustness to distribution shift. The performance degrades systematically across all models, shift types, and pre-training strategies, and the relationship between ID and OOD performance documented for traditional models extends to TFMs. Pre-training corpus in TFMs does not provide OOD robustness. Although models pre-trained on real-world data achieve superior OOD performance, they show shift gaps similar to those of synthetic-prior models, suggesting that their advantage is driven by stronger ID fitting rather than an inherent capacity for OOD generalization. This work extends OOD benchmarks for tabular data and provides evidence supporting their application in high-stakes scenarios involving structural distribution shifts.

\begin{credits}
\subsubsection{\ackname} 
The authors acknowledge the Sigma AI Lab at Universidad San Francisco de Quito (USFQ) for providing the computational resources and AI infrastructure that enabled this research. Also, this work was supported through the Poli-Grants Program under Grant Numbers 41990 and 39820. 

\end{credits}
%
%
%

\bibliographystyle{splncs04}
\bibliography{bibliography}

@article{behre2026context,
  title={In-Context Forecasting in Supply Chains: Evaluating the Promise and Limits of Tabular Foundation Models},
  author={Behre, Ole},
  year={2026}
}

@article{zhang2026mitra,
  title={Mitra: Mixed synthetic priors for enhancing tabular foundation models},
  author={Zhang, Xiyuan and Maddix Robinson, Danielle and Yin, Junming and Erickson, Nick and Ansari, Abdul Fatir and Han, Boran and Zhang, Shuai and Akoglu, Leman and Faloutsos, Christos and Mahoney, Michael and others},
  journal={Advances in neural information processing systems},
  volume={38},
  pages={15795--15840},
  year={2026}
}

@article{grinsztajn2026tabpfn,
  title={Tabpfn-3: Technical report},
  author={Grinsztajn, L{\'e}o and Fl{\"o}ge, Klemens and Key, Oscar and Birkel, Felix and Jund, Philipp and Roof, Brendan and Manium, Mihir and Hoo, Shi Bin and B{\"u}hler, Magnus and Garg, Anurag and others},
  journal={arXiv preprint arXiv:2605.13986},
  year={2026}
}

@article{jiang2026representation,
  title={Representation learning for tabular data: A comprehensive survey},
  author={Jiang, Jun-Peng and Liu, Si-Yang and Cai, Hao-Run and Zhou, Qi-Le and Ye, Han-Jia},
  journal={IEEE Transactions on Pattern Analysis and Machine Intelligence},
  year={2026},
  publisher={IEEE}
}

@article{gardner2023benchmarking,
  title={Benchmarking distribution shift in tabular data with tableshift},
  author={Gardner, Josh and Popovic, Zoran and Schmidt, Ludwig},
  journal={Advances in Neural Information Processing Systems},
  volume={36},
  pages={53385--53432},
  year={2023}
}

@article{lee2026multitabpfn,
  title={MultiTabPFN: Codebook-based extensions of TabPFN for high-class-count tabular classification},
  author={Lee, Kyungeun},
  journal={Neural Networks},
  pages={108932},
  year={2026},
  publisher={Elsevier}
}

@article{ruiz2024tabpfn,
  title={A TabPFN-based intrusion detection system for the industrial internet of things: S. Ruiz-Villafranca et al.},
  author={Ruiz-Villafranca, Sergio and Rold{\'a}n-G{\'o}mez, Jos{\'e} and G{\'o}mez, Juan Manuel Castelo and Carrillo-Mond{\'e}jar, Javier and Martinez, Jos{\'e} Luis},
  journal={The Journal of Supercomputing},
  volume={80},
  number={14},
  pages={20080--20117},
  year={2024},
  publisher={Springer}
}

@article{ruiz2025wfe,
  title={WFE-Tab: Overcoming limitations of TabPFN in IIoT-MEC environments with a weighted fusion ensemble-TabPFN model for improved IDS performance},
  author={Ruiz-Villafranca, Sergio and Rold{\'a}n-G{\'o}mez, Jos{\'e} and Carrillo-Mondejar, Javier and Martinez, Jos{\'e} Luis and Ga{\~n}{\'a}n, Carlos H},
  journal={Future Generation Computer Systems},
  volume={166},
  pages={107707},
  year={2025},
  publisher={Elsevier}
}

@article{zhang2025limix,
  title={Limix: Unleashing structured-data modeling capability for generalist intelligence},
  author={Zhang, Xingxuan and Ren, Gang and Yu, Han and Yuan, Hao and Wang, Hui and Li, Jiansheng and Wu, Jiayun and Mo, Lang and Mao, Li and Hao, Mingchao and others},
  journal={arXiv preprint arXiv:2509.03505},
  year={2025}
}

@article{hollmann2025accurate,
  title={Accurate predictions on small data with a tabular foundation model},
  author={Hollmann, Noah and M{\"u}ller, Samuel and Purucker, Lennart and Krishnakumar, Arjun and K{\"o}rfer, Max and Hoo, Shi Bin and Schirrmeister, Robin Tibor and Hutter, Frank},
  journal={Nature},
  volume={637},
  number={8045},
  pages={319--326},
  year={2025},
  publisher={Nature Publishing Group UK London}
}

@article{somvanshi2026survey,
  title={A survey on tabular data: from tree-based methods to tabular deep learning},
  author={Somvanshi, Shriyank and Das, Subasish and Javed, Syed and Antariksa, Gian and Hossain, Ahmed},
  journal={ACM Computing Surveys},
  year={2026},
  publisher={ACM New York, NY}
}

@article{somepalli2021saint,
  title={Saint: Improved neural networks for tabular data via row attention and contrastive pre-training},
  author={Somepalli, Gowthami and Goldblum, Micah and Schwarzschild, Avi and Bruss, C Bayan and Goldstein, Tom},
  journal={arXiv preprint arXiv:2106.01342},
  year={2021}
}

@inproceedings{arik2021tabnet,
  title={Tabnet: Attentive interpretable tabular learning},
  author={Arik, Sercan {\"O} and Pfister, Tomas},
  booktitle={Proceedings of the AAAI conference on artificial intelligence},
  volume={35},
  number={8},
  pages={6679--6687},
  year={2021}
}

@article{ma2024tabdpt,
  title={Tabdpt: Scaling tabular foundation models on real data},
  author={Ma, Junwei and Thomas, Valentin and Hosseinzadeh, Rasa and Labach, Alex and Kamkari, Hamidreza and Cresswell, Jesse C and Golestan, Keyvan and Yu, Guangwei and Caterini, Anthony L and Volkovs, Maksims},
  journal={arXiv preprint arXiv:2410.18164},
  year={2024}
}

@article{lara2025interpretable,
  title={Interpretable and multimodal fusion methodology to predict severe hypoglycemia in adults with type 1 diabetes},
  author={Lara-Abelenda, Francisco J and Chushig-Muzo, David and W{\"a}gner, Ana M and Tayefi, Maryam and Soguero-Ruiz, Cristina},
  journal={Engineering Applications of Artificial Intelligence},
  volume={144},
  pages={110142},
  year={2025},
  publisher={Elsevier}
}

@article{qu2026tabiclv2,
  title={TabICLv2: A better, faster, scalable, and open tabular foundation model},
  author={Qu, Jingang and Holzm{\"u}ller, David and Varoquaux, Ga{\"e}l and Morvan, Marine Le},
  journal={arXiv preprint arXiv:2602.11139},
  year={2026}
}

@article{yang2024generalized,
  title={Generalized out-of-distribution detection: A survey},
  author={Yang, Jingkang and Zhou, Kaiyang and Li, Yixuan and Liu, Ziwei},
  journal={International Journal of Computer Vision},
  volume={132},
  number={12},
  pages={5635--5662},
  year={2024},
  publisher={Springer}
}

@article{chushig2024characterizing,
  title={Characterizing the impact of physical activity on patients with Type 1 Diabetes using statistical and machine learning models},
  author={Chushig-Muzo, David and Calero-D{\'\i}az, Hugo and Fabelo, Himar and {\AA}rsand, Eirik and van Dijk, Peter Ruben and Soguero-Ruiz, Cristina},
  journal={Applied Sciences},
  volume={14},
  number={21},
  pages={9870},
  year={2024},
  publisher={MDPI}
}

\end{document}